%% file: eacl23-topic-ontologies-for-argumentation-frame.tex
\pdfoutput=1
\documentclass[11pt]{article}
\usepackage{eacl2023}
\usepackage{hyperref}
\PassOptionsToPackage{hyphens}{url}
\usepackage[american]{babel}
\usepackage[bottom]{footmisc}
\usepackage[pdftex]{graphicx}

\usepackage{times}
\usepackage{type1cm}
\usepackage[T1]{fontenc}
\usepackage[latin1]{inputenc}
\usepackage{microtype}
\usepackage{xcolor}

\newcommand{\Ni}{(1)~}
\newcommand{\Nii}{(2)~}
\newcommand{\Niii}{(3)~}

\newcommand{\bslabel}[1]{\smallskip\noindent{\bfseries #1}}

\usepackage{url}
\RequirePackage{color}
\RequirePackage{soul}
\setstcolor{blue}

\usepackage{booktabs}

\DeclareGraphicsExtensions{.pdf,.ai,.jpg,.png}
\setkeys{Gin}{pagebox=artbox}

\usepackage{etoolbox}
\makeatletter
\patchcmd\@combinedblfloats{\box\@outputbox}{\unvbox\@outputbox}{}{%
}%
\makeatother

\begin{document}
\input{eacl23-topic-ontologies-for-argumentation-pre}
\input{eacl23-topic-ontologies-for-argumentation-part1}
\input{eacl23-topic-ontologies-for-argumentation-part2}
\input{eacl23-topic-ontologies-for-argumentation-part3}
\input{eacl23-topic-ontologies-for-argumentation-part4}
\input{eacl23-topic-ontologies-for-argumentation-part5}
\input{eacl23-topic-ontologies-for-argumentation-part6}
\input{eacl23-topic-ontologies-for-argumentation-sum}

\bibliography{eacl23-topic-ontologies-for-argumentation-lit}
\newpage
\input{eacl23-topic-ontologies-for-argumentation-limitations}
\bibliographystyle{acl_natbib}

\input{eacl23-topic-ontologies-for-argumentation-supplementary.tex}

\end{document}

%% file: eacl23-topic-ontologies-for-argumentation-pre.tex
\setlength\titlebox{8cm}
\title{Topic Ontologies for Arguments}

\author{%
	Yamen Ajjour \\
	Lebiniz University Hannover\\
	\texttt{y.ajjour@ai.uni-hannover.de} \hspace*{4ex}\\%
	\And
	Johannes Kiesel \\
	Bauhaus-Universit\"at Weimar \\
	\texttt{johannes.kiesel@uni-weimar.de} \\%
	\AND
	Benno Stein \\
	Bauhaus-Universit\"at Weimar \\
	\texttt{benno.stein@uni-weimar.de} \hspace*{4ex}\\%
	\And
	Martin Potthast \\
	Leipzig University \\
	\texttt{martin.potthast@uni-leipzig.de} \\%
}

\date{}

\maketitle

\begin{abstract}
Many computational argumentation tasks, like stance classification, are topic-dependent: the effectiveness of approaches to these tasks significantly depends on whether the approaches were trained on arguments from the same topics as those they are tested on. So, which are these topics that researchers train approaches on? This paper contributes the first comprehensive survey of topic coverage, assessing 45~argument corpora. For the assessment, we take the first step towards building an argument topic ontology, consulting three diverse authoritative sources: the World Economic Forum, the Wikipedia list of controversial topics, and Debatepedia. Comparing the topic sets between the authoritative sources and corpora, our analysis shows that the corpora topics---which are mostly those frequently discussed in public online fora---are covered well by the sources. However, other topics from the sources are less extensively covered by the corpora of today, revealing interesting future directions for corpus construction.
\end{abstract}

%% file: eacl23-topic-ontologies-for-argumentation-part1.tex
\section{Introduction}

The term ``topic'' refers to a text's subject matter. A text can be about one or more topics; the relation underlying topics and texts is called ``aboutness'' \cite{yablo:2014}. Topics play a central role in argumentation, since they constrain or guide strategies and rhetorical devices by providing the accepted and expected universe of discourse. Also, the view of pragma-dialectics in argumentation emphasizes that argumentation is topic-dependent \cite{vaneemeren:2015}: ``The basic aspects of strategic maneuvering [\ldots] are making an expedient selection from the `topical potential' available at a certain discussion.'' Though debaters often use commonplace arguments across topics \newcite{bilu:2019}, this is only possible for related topics: a black-market argument, for example, applies to topics like banning drugs or banning guns. When developing computational models to extract, analyze, or generate arguments, however, one should thus ensure a wide topic coverage in model training to improve the model's generalizability (e.g., as recently shown by \citealp{reuver:2021}).

A set of topics may be organized as a graph, sometimes called ``topic space''. Information theorists and library scientists map hierarchical topic relations within ontologies \cite{hjorland:2001}. Here, topics are labeled with a subject heading, i.e., a phrase from a controlled vocabulary which concisely and discriminatively describes a topic. Library ontologies are not designed with argumentation tasks in mind, but other ontology efforts specifically address {\em argumentative} topic spaces. We identified and harnessed three authoritative sources of ontologic knowledge that cover global issues, controversies, and popular debates: the World Economic Forum's ``Strategic Intelligence'' site, Wikipedia's list of controversial topics, and Debatepedia's debate classification system (cf.~Section~\ref{topic-ontologies}).

We contribute a comprehensive overview of argument corpora and their topic coverage as per the mentioned ontologies. The coverage of corpora that provide topic labels is manually assessed by aligning each label to the ontologies' topics, computing the proportion of ontology topics covered by a corpus, and the distribution of corpus arguments in an ontology. Our analyses show that existing corpora focus on a subset of possible topics (cf.~Section~\ref{topic-coverage-corpus-topic-labels}). For the corpora without topic labels, we categorize their argumentative texts by measuring the semantic relatedness of corpus documents to ontology topics. Given the 748~topic, this is a challenging classification, for which we achieve a remarkable F$_1$ of~0.59 (cf.~Section~\ref{topic-coverage-unit-categorization}).%
\footnote{Anonymized data at \url{https://zenodo.org/record/3928096}.}

Altogether, we lay the foundation for the study and systematic exploration of controversial topics within computational argumentation analysis. The identified authoritative resources already capture quite comprehensively their respective domains. Future work will have to extend our approach to other topic spaces, such as business, domestic, historic, and scientific argumentation spaces.

%% file: eacl23-topic-ontologies-for-argumentation-part2.tex
\section{Related Work}
\label{related-work}

Our review of related work focuses on the role of the variable ``topic'' in computational argumentation. Moreover, we briefly review topic ontologies and hierarchical topic classification.

\subsection{Topics in Computational Argumentation}


In computational argumentation, arguments are typically modeled as compositions of argument units, where an argument unit is represented as a span of text. \newcite{habernal:2016c} adopts \newcite{toulmin:1958}'s~(1958) model, which defines six unit types, among which are ``claim'' and ``data''. \newcite{wachsmuth:2017b} employ a more basic model of two units, which defines an argument as a claim or conclusion supported by one or more premises. These models capture arguments without explicitly identifying the topic they address. \newcite{levy:2014} consider claims to be topic-dependent and study their detection in the context of a random selection of 32~topics from \url{idebate.org}. This work raises the question why topic-dependence has not been addressed more urgently until now.


Key tasks for computational argumentation include the mining of arguments from natural language \cite{moens:2007,stein:2016g}, classifying their stances with regard to a thesis \cite{bar-haim:2017}, and analyzing which arguments are more persuasive \cite{tan:2016,habernal:2016c}. Current approaches to these tasks rely on supervised classification.
\newcite{daxenberger:2017} show that supervised classifiers fail to generalize across domains ($\sim$ topics). More recently, \newcite{stab:2018a} tweak Bi\-LSTM \cite{graves:2005} to integrate the topic while jointly detecting (1)~whether a sentence is an argument and (2)~its stance to the topic. The designed neural network outperforms BiLSTM without topic integration in both tasks; the approach gives further evidence for the topic-dependence of argument mining and stance classification. Whether model transfer between more closely related topics works better is unknown. As a first step, \newcite{reuver:2021} show that cross-topic stance-classification with BERT~\cite{devlin:2018} produces mixed results depending on the topics, but misses the relations between the topics. \newcite{gu:2018} show that integrating the topic of an argument helps assessing its persuasiveness.

Topic plays a central role in argument retrieval and generation since it defines what arguments are relevant. Argument retrieval aims at delivering pro and con arguments on a given topic query. A major challenge in argument retrieval is the grouping of arguments that address common aspects of a topic. As shown by~\newcite{reimers:2019} and \newcite{stein:2019z}, integrating the topic is an important step while clustering arguments. For argument generation, \newcite{bilu:2019} introduce an approach that matches an input topic against a list of topics that are paired with sets of topic-adjustable commonplace arguments (e.g.,~black-market arguments). In a similar vein, \newcite{bar-haim:2019} identify consistent and contrastive topics for a given topic with the goal of expanding the topic in a new direction (e.g.,~fast food versus obesity). Both approaches show the merit of utilizing argument topic ontologies in argument generation.
Perhaps only abstract argumentation can be conceived of as topic independent, since it studies the structure and relations among arguments more than their language.

\subsection{Topic Ontologies}

In information science, an ontology is defined as ``an explicit specification of a conceptualization'' \cite{gruber:1993}. Topic ontologies are a specific type of ontologies which specify topics as nodes of a directed acyclic graph. An edge in the graph then implies an ``is part of''-relation between the topics \cite{xamena:2017}. The effort in creating topic ontologies ranges from ad-hoc decisions (e.g.,~tags for blog posts) to extensive classification schemes for libraries. The oldest classification scheme that is still used today in libraries is the Dewey Decimal Classification.
It has been translated into over 30~languages, and it contains several tens of thousands of classes. Most topic ontologies focus on a specific domain, such as a the ACM Computing Classification System for computer science, or DMOZ for web pages.%
\footnote{\url{https://dl.acm.org/ccs}\ \ and\ \ \url{https://dmoz-odp.org/}}
The only topic ontology directly linked to arguments is that of Debatepedia.


\subsection{Hierarchical Text Classification}

Hierarchical text classification aims at classifying a document into a class hierarchy. Depending on how the hierarchical structure is exploited, classification can be done top-down (from higher classes downwards), bottom-up, or flat (ignoring hierarchical relations) \cite{silla:2011}. Researchers usually train supervised classifiers for each class in the hierarchy \cite{sun:2001}.

%% file: eacl23-topic-ontologies-for-argumentation-part3.tex
\section{Survey of Argument Corpora}
\label{survey}

\input{table-argument-corpora}

To study arguments and computational argumentation tasks, researchers compile corpora with argumentative texts. To the best of our knowledge, Table~\ref{table-argument-corpora} lists all corpora dedicated to \mbox{argumentation} to 2020. We review these corpora and their associated publications with regard to what are the sources of arguments, what is the granularity of the corpus, what is the size of the corpora in terms of their units, and which and how many different topics are covered in them. Reviewing all papers citing a corpus, we also analyzed how many experiments were carried out using them.

The most elaborate discussion of topic selection is given in \newcite{habernal:2016c}, who chose six topics (homeschooling, public versus private schools, redshirting, prayers in schools, single sex education, mainstreaming) to focus on different education-related aspects. The broadest selection of topics is reported by the researchers of IBM Debater,%
\footnote{\fontsize{7.5pt}{9pt}\selectfont\url{https://www.research.ibm.com/haifa/dept/vst/debating_data.shtml}}
who obtain arguments from Wikipedia.
The only other work mentioning their source of topics stems from \newcite{stab:2018a}, who randomly select 8~topics from two lists of controversial topics that originate from an online library and the debate portal ProCon.org, respectively. \newcite{peldszus:2015} predefine a set of topics and give writers the freedom to choose which one to write about, but nothing is said about where the set of predefined topics originate from. \newcite{conard:2012} and \newcite{hasan:2014}  explicitly select topics (1~and~4, respectively). For all other corpora with topic labels, their authors do not argue on choosing topics, nor selection or sampling criteria. Neither do the authors of corpora without topic labels.

\begin{figure*}
\centering
\includegraphics{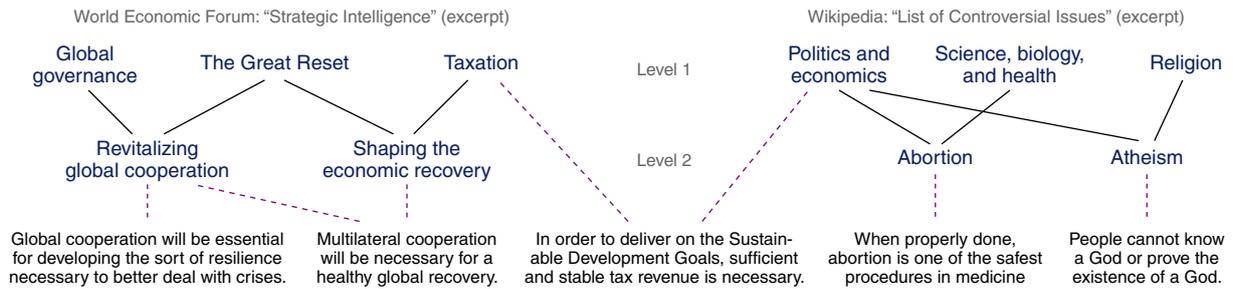}
\caption{Example for an assignment of arguments (bottom) to topics of a two-leveled ontology. Level~2 topics are subtopics of their linked Level~1 topics. Arguments linked to a Level~2 topic also pertain to its Level~1 upper-topics.}
\label{topic-ontology-example3}
\end{figure*}

Altogether, it appears that the best practices in argumentation do not as of yet consider topic sampling as a prerequisite task to ensure coverage of a certain domain of interest, and diversity. Based on our review, we presume three basic topic selection directives are in use today:
\Ni
{\em Manual selection.}
Topics are manually defined or selected. Although the process may be random, when aiming for controversial topics, one may often end up with commonplace topics in Western culture (e.g., abortion, death penalty, gay marriage), despite them them being still relevant and important today.
\Nii
{\em Source-driven (greedy within a time-span).}
A source of argument ground truth is either exploited in its entirety, or a maximum subset fulfilling desired properties is used. Since argument-related ground truth is hard to come by, it is understandable that all available sources are being exploited.
\Niii
{\em Source-driven (sampled).}
A source or argument ground truth is exploited and a subset is sampled. Here, it may be infeasible to exploit a source in its entirety. \newcite{stein:2016s} randomly select 300~documents from three websites. \newcite{park:2018} and \newcite{stab:2017} do not mention anything about their sampling process.
In general, both source-driven corpus construction approaches inevitably incurs the source's idiosyncracies of topic selection, both in terms of skew towards certain topics. Scaling up may or may not be a remedy for this problem.

We assess how many experiments have been reported on each of the corpora by collecting the publications referring to a corpus as per Google Scholar, focusing on conference and journal papers, but excluding books and web pages. We then check whether the cited corpus is mentioned in its data, experiment, or results section. As can be seen in Table~\ref{table-argument-corpora}, corpora with fewer topics tend to be used more often in experiments than those with larger amounts. In total, 82~experiments were carried out on argument corpora with no clearly defined topic selection directive. The skew towards smaller-scale experiments may affect generalizability.

%% file: table-argument-corpora.tex
\begin{table*}
\centering
\footnotesize
\renewcommand{\arraystretch}{1.2}
\setlength{\tabcolsep}{1.7pt}
\begin{tabular}{@{}llllrrr@{}}

\toprule
\bfseries Corpus & \bfseries Authors & \bfseries Source & \bfseries Unit granularity & \bfseries Units & \bfseries Topics & \bfseries Exp. \\
\midrule
\multicolumn{7}{@{}c@{}}{\bfseries Manual selection} \\
Arguing Subjectivity             & \newcite{conard:2012}      & Editorials          & Editorial/blog      &                  84 &               1 &  1 \\
Argumentative Sentences          & \newcite{shnarch:2020}     & Wikipedia           & Arguments           &                 700 &              20 &  1 \\
Argument Facet Similarity        & \newcite{misra:2016}       & Debate portals      & Argument            &               6,188 &               3 &  8 \\
Claim and Evidence 1             & \newcite{aharoni:2014}     & Wikipedia           & Wikipedia article   &                 315 &              33 & 18 \\
Claim and Evidence 2             & \newcite{rinott:2015}      & Wikipedia           & Wikipedia article   &                 547 &              58 & 12 \\
Claim Generation                 & \newcite{gretz:2020}       & Generated text      & Argument Unit       &               2,839 &             136 &  1 \\
Claim Stance                     & \newcite{bar-haim:2017}    & Wikipedia           & Argument Unit       &               2,394 &              55 & 10 \\   
Claim Sentence Search            & \newcite{levy:2018}        & Wikipedia           & Argument unit       & \kern-1em 1,492,077 &             150 &  3 \\
COMARG                           & \newcite{boltuzic:2014}    & Debate portals      & Argument pair       &               2,298 &                2&  3 \\
Evidence Sentences               & \newcite{schnarch:2018}    & Wikipedia           & Argument unit       &               5,783 &             118 &  5 \\
Evidence Sentences 2             & \newcite{eindor:2020}      & Wikipedia           & Argument unit       &              29,429 &             221 &  3 \\
Evidence Quality                 & \newcite{gleize:2019}      & Wikipedia           & Argument pair       &               5,697 &              69 &  1 \\
ICLE Essay Scoring               & \newcite{persing:2010}     & Essays              & Essay               &               1,000 &              10 & 11 \\
Ideological Debates Reasons      & \newcite{hasan:2014}       & Debate portals      & Argument            &               4,903 &               4 & 10 \\
Internet Argument Corpus v2      & \newcite{abbott:2016}      & Web                 & Discussion          &              16,555 &              19 & 18 \\
Key Point Analysis               & \newcite{barhaim:2020}     & Wikipedia           & Argument            &              24,093 &              28 &  2 \\
Micro Text v1                    & \newcite{peldszus:2015}    & Essays              & Essay               &                 112 &              18 &  7 \\
Micro Text v2                    & \newcite{skeppstedt:2018}  & Essays              & Essay               &                 171 &              35 &  1 \\
Multilingual Argument Mining     & \newcite{toledo-ronen:2020}& Wikipedia           & Argument unit       &              65,708 &             347 &  2 \\
Political Argumentation          & \newcite{menini:2018}      & Presidential debate & Argument pair       &               1,462 &               5 &  3 \\
Record Debating Dataset 2        & \newcite{mirkin:2018}      & Debating            & Speech              &                 200 &              50 &  5 \\
Record Debating Dataset 3        & \newcite{lavee:2019}       & Debating            & Speech              &                 400 &             199 &  1 \\
Record Debating Dataset 4        & \newcite{orbach:2019}      & Debating            & Speech              &                 200 &             50  &  1 \\
Record Debating Dataset 5        & \newcite{orbach:2020}      & Debating            & Speech              &               3,562 &             397 &  1 \\
Sci-arg                          & \newcite{lauscher:2018}    & Scientific papers   & Paper               &                  40 &                1&  3 \\
UKP Sentential                   & \newcite{stab:2018a}       & Web                 & Argument            &              25,492 &               8 & 13 \\ 
UKP Aspect                       & \newcite{reimers:2019}     & Web                 & Argument pair       &               3,595 &              28 &  3 \\
UKPConvArg1                      & \newcite{habernal:2016a}   & Debate portals      & Argument pair       &              11,650 &              16 & 10 \\
UKPConvArg2                      & \newcite{habernal:2016b}   & Debate portals      & Argument pair       &               9,111 &              16 &  3 \\
WebDiscourse                     & \newcite{habernal:2016c}   & Web                 & Document            &                 340 &               6 &  7 \\
Webis-debate-16                  & \newcite{alkhatib:2016a}   & Debate portals      & Debate              &                 445 &              14 &  3 \\
\midrule
\multicolumn{7}{@{}c@{}}{\bfseries Source-driven: greedy within a time-span} \\
AIFdb                            & \newcite{bex:2013}         & Web                 & Argument unit       &              67,408 & \color{gray}n/a &  7 \\  
Args-me                          & \newcite{stein:2019p}      & Debate portals      & Argument            &             387,692 & \color{gray}n/a &  3 \\
ChangeMyView                     & \newcite{tan:2016}         & Discussion forum    & Post/comment        &              14,066 & \color{gray}n/a & 21 \\
DebateSum                        & \newcite{roush:2020}       & Debating            & Debate              &             187,386 & \color{gray}n/a &  1 \\
Intelligence Squared Debates     & \newcite{zhang:2016}       & Debate portals      & Debate              &                 108 & \color{gray}n/a &  3 \\
Kialo                            & \newcite{kialo}            & Debate portals      & Argument unit       &             331,684 & \color{gray}n/a &  3 \\   
Political Speech                 & \newcite{lippi:2016}       & Ministerial debate  & Argument unit       &                 152 & \color{gray}n/a &  1 \\
USElecDeb60To16                  & \newcite{haddadan:2019}    & Presidential debate & Debate              &                  42 & \color{gray}n/a &  1 \\
\midrule
\multicolumn{7}{@{}c@{}}{\bfseries Source-driven: sampled} \\                                                                                             
Argument Annotated Essays        & \newcite{stab:2017}        & Essays              & Essay               &                 402 & \color{gray}n/a & 28 \\
E-rulemaking                     & \newcite{park:2018}        & Discussion forum    & Argument            &                 731 & \color{gray}n/a &  3 \\
ECHR                             & \newcite{poudyal:2020}     & Law Case            & Argument            &                 743 & \color{gray}n/a &  1 \\
Editorials                       & \newcite{stein:2016s}      & Editorials          & Editorial           &                 300 & \color{gray}n/a &  8 \\
GAQCorpus                        & \newcite{ng:2020}          & Web                 & Argument            &               6,424 & \color{gray}n/a &  1 \\
IDebate Persuasiveness           & \newcite{persing:2017}     & Debate portals      & Argument            &               1,205 & \color{gray}n/a &  1 \\

\bottomrule
\end{tabular}
\caption{Survey of argument corpora indicating data source, unit granularity, and size in terms of units and topics (if authors remarked on it). The unit granularity is the one in the corpus' files, using premises and conclusions as one unit each and the best context-preserving unit for corpora featuring multiple granularities. We presume these topic selection directives from the corpus description: either {\em manual selection} by the authors, or {\em source-driven}---i.e., the topics in the selected source(s)---from the units of a specific {\em time-span} or by random {\em sampling}. Experiments (Exp.) denotes the count of papers that use the corpus in an experiment among those papers that cite the corpus' paper.}
\label{table-argument-corpora}
\end{table*}

%% file: eacl23-topic-ontologies-for-argumentation-part4.tex
\section{Acquiring Argument Topic Ontologies}
\label{topic-ontologies}

Topic ontologies provide for a knowledge organization principle, and, especially if widely accepted, also a standard. They are typically modeled as directed acyclic graphs, where nodes correspond to topics and edges indicate ``is part of'' relations; topics that are part of other topics are called their subtopics. A topic ontology is often displayed in levels, starting with the topics that are not subtopics of others, continuing recursively with each lower level of subtopics. Figure~\ref{topic-ontology-example3} shows an excerpt of a two-level topic ontology for arguments.

The identification of the topics to be included in an argument topic ontology, as well as their relations, requires domain expertise. Building an all-encompassing ontology thus requires experts from every top-level domain where argumentation of scientific interest is expected. In the following, we suggest and outline three authoritative sources of relevant topic ontologies, which comprise a wide selection of important argumentative topics.


\bigskip
\bslabel{World Economic Forum (WEF)}
The World Economic Forum is a not-for-profit foundation that coordinates organizations from both the public and the private sector to work on economical and societal issues. As part of their efforts, their ``Strategic Intelligence'' platform%
\footnote{\fontsize{7.7pt}{9pt}\selectfont\url{https://intelligence.weforum.org}}
strives to inform decision makers on domestic and global topics, specifically global issues (e.g., artificial intelligence and climate change), industries (e.g., healthcare delivery and private investors), and economies (e.g., Africa and ASEAN). Domain experts for each topic curate a stream of relevant news articles which they each tag with 4-9 subtopics of their topic (e.g., the continuous monitoring of mental health).



\bslabel{Wikipedia}
Wikipedia strives for a neutral point of view, but many topics of public interest are discussed controversially. Some editors thus curate a list of such controversial articles to highlight where special care is needed, grouped into 14~top-level topics (e.g., environment and philosophy) and \mbox{4-176}~subtopics (e.g., creationism and pollution).%
\footnote{\fontsize{6.5pt}{9pt}\selectfont\url{https://en.wikipedia.org/wiki/Wikipedia:List_of_controversial_issues}}
Omitted is the ``People'' topic and articles on countries; their controversiality is not universal.

\enlargethispage{0.25\baselineskip}
\bslabel{Debatepedia}
The debate portal's goal is to create an encyclopedia of debates which are organized as ``pro'' and ``con'' arguments. A list of 89~topics helps visitors to browse the debates. The debates are contributed by anonymous web users, which makes the covered topics easily accessible. Topics in Debatepedia tend to address issues of Western culture. For example, the topic ``United States'' covers 306 debates while ``Third World'' covers 12 debates. The project is no longer maintained, but can be accessed through the Wayback Machine.\footnote{\fontsize{6.5pt}{9pt}\selectfont\url{https://web.archive.org/web/20180222051626/http://www.debatepedia.org/en/index.php/Welcome_to_Debatepedia\%21}}

The three ontologies are publicly accessible, and two of them are actively maintained and updated. Acquiring the ontologies is straightforward---not straightforward is to make use of them. A key task associated with every topic ontology is to categorize a given document. Having just a short string label describing a (potentially multifaceted) topic, such as ``The Great Reset'', renders this task exceedingly difficult. Fortunately, domain experts have been pre-categorizing documents into the aforementioned ontologies. In particular, regarding the WEF, invited domain experts categorize news articles for every topic, regarding Wikipedia, the text of the associated wiki articles is available, as are the associated debates for Debatepedia.

Articles that are categorized into Level~2 topics are propagated up to their respective Level~1 topics. Table~\ref{table-topic-stats-and-classification} shows the large differences between the ontologies. The WEF ontology contains the most topics and links the most documents, which contain the most tokens overall. The topics at Wikipedia Level~2 are just linked to a single article each, so every topic's amount of text is smaller. The number of authors reflects the number of editors.


%% file: eacl23-topic-ontologies-for-argumentation-part5.tex
\section{Topic Coverage}
\label{topic-coverage-corpus-topic-labels}

To assess the topic coverage of the argument corpora in light of the three ontologies, we map the topic labels of those corpora providing them to their matching ontology topics.


\begin{figure*}[t]
\centering
\includegraphics{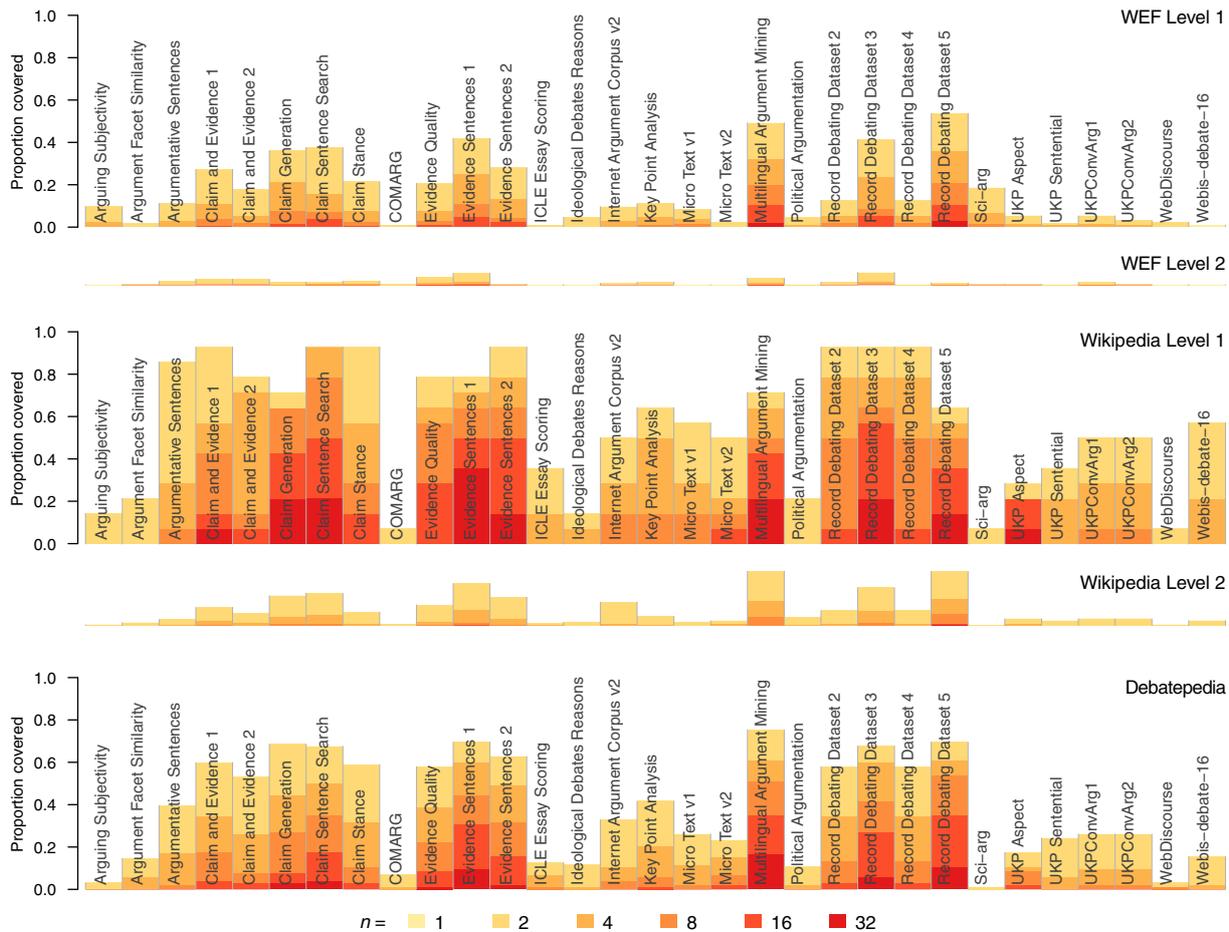}
\caption{Proportion of ontology topics covered by at least $n$~corpus topics (per ontology level and per corpus).}
\label{ontology-topics-covered-per-corpus}
\end{figure*}

\subsection{Topic Label Normalization}

Table~\ref{table-argument-corpora} lists 31~argument corpora that provide topic labels. Altogether 2,117~different labels have been assigned. They are concise descriptions of the main issues of an argument and have been provided by the corpus authors. The labels follow the style of the text register of the respective corpus: In essays, for instance, topics are usually thesis statements, while Wikipedia-derived corpora use article titles, and the topics of debate corpora include clich{\'e}s such as ``This house should''. Often, topic labels express a stance towards a target issue, e.g., ``ban guns''. Five types of topic labels can be distinguished: concept, comparison of concepts, conclusion (includes claim and thesis), question, and imperative. We normalize the topic labels by converting all concepts to singular form, removing clich{\'e}s, and dropping stance-indicating words such as ``legalize''. Our normalization aims at retaining only the central target issue of a topic label and leads to 748~unique topic labels.

\subsection{Mapping Topic Labels to Ontology Topics}

Using the preprocessed topic labels as queries, we retrieve for each topic label the~50 top-most relevant topics in each level of the three ontologies. To facilitate the retrieval of ontology topics, we employ a BM25-weighted \cite{robertson:2004} index of the concatenated documents for each topic. BM25 is a widely used modified version of TF-IDF \cite{croft:2009}. This enables us to narrow down the mapping of a topic label to a manageable size. Except for a handful of cases, 50~ontology topics can be retrieved for each topic label. The topic labels were then manually mapped to an ontology topic, if they form synonyms, or if the former is a subtopic of the latter---which thus indicates that all arguments in the corpus with that topic label are about the ontology topic. A topic label can thus be mapped to multiple ontology topics. For example, the topic label ``plastic bottles'' is mapped to ``pollution'' and ``recycling'' in Wikipedia Level~2.


\begin{figure*}[t]
\centering
\includegraphics{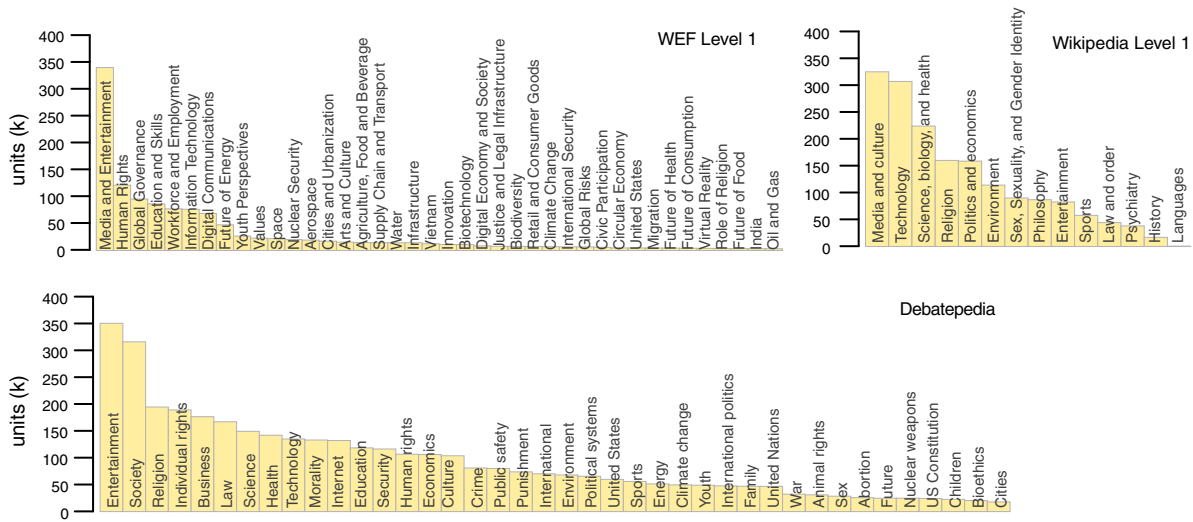}
\caption{Distribution of corpora units over the top matching topics in an ontology (31 labeled corpora).}
\label{histogram-ontology-topics-in-corpora}
\end{figure*}

\subsection{Analysis of Topic Coverage}

Table~\ref{table-topic-stats-and-classification} shows general statistics of this mapping of corpora topic labels to ontology topics. Most of the topic labels (2,002 out of 2,117) are mapped to at least one Debatepedia topic while only 355~labels are mapped to WEF Level~2 topics. For Wikipedia Level~2, only 285 out of the 748~topics are actually covered by argument corpora. Already this first analysis suggests that existing argument corpora cover typically a small subset of possible argumentative topics that people are trained to debate. For those topic labels that can be mapped are mapped on average to 2.8~topics in Debatepedia, to 1.24~topics in Wikipedia Level~1, and to 1.48~topics in WEF Level~1. As discussed in Section~\ref{topic-ontologies}, topics in Debatepedia focus on the Western culture and are easily accessible, whereas topics in WEF require deeper domain knowledge and have more global relevance. The broad coverage of Debatepedia's topics indicates that the studied argument corpora focus on common topics that are easily approachable while global issues or those that need domain knowledge lack coverage. 

For a more fine-grained analysis, Figure~\ref{ontology-topics-covered-per-corpus} illustrates the differences regarding the number of ontology topics covered by a corpus: while topics in Wikipedia Level~1 are covered well by some argument corpora, topics in Wikipedia and WEF Level~2 are covered only marginally. Note that topic coverage varies significantly between the corpora: the Claim Sentence Search dataset's topics cover 93\% of the Wikipedia Level~1 topics, while the Ideological Debates Reasons dataset covers only~14\%. The colors show the topic granularity of the corpus; especially the Record Debating Dataset~3 dataset is fine-grained: as the highest value, 36~of its topics are mapped to the Wikipedia Level~1 category ``Politics and Economics''.

Figure~\ref{histogram-ontology-topics-in-corpora} shows how the set of the units of the 31 labeled corpora distribute over the top matching topics in Debatepedia, Wikipedia Level~1, and WEF Level~1. Distributions over Level~2 are omitted for brevity and can be found in the supplementary material. The distribution is significantly skewed: while the top ten topics in Debatepdia are matched by 340k to 150k corpora units, the top ten topics in WEF Level~1 are matched by 340k to 20k corpora units. The comparison between the three ontologies supports our previous finding that argument corpora cover easily accessible topics (e.g., ``Media and Entertainment'' and ``Society'').

%% file: eacl23-topic-ontologies-for-argumentation-part6.tex
\section{Unit Categorization}
\label{topic-coverage-unit-categorization}

\input{table-topic-stats-and-classification}

The previous analysis is done on those argument corpora which contain topic labels. About a third of the argument corpora are thus excluded from that analysis. As a step toward assessing their topic coverage, we map the ontology topics for a unit (cf.\ Table~\ref{table-argument-corpora}) in an argument corpus by treating the unit as a (long) query in a standard information retrieval setup, where ontology topics are the retrieval targets. The documents categorized into each topic have been concatenated and used as the topic's representation. Though the documents associated with a topic are not necessarily argumentative, they cover the salient aspects of the topic. 

To retrieve topics for a corpus unit, we implement and evaluate the following approaches:  Semantic Interpretation (SI) and SI with Text Embeddings (T2V-SI). The Semantic interpretation approach computes the semantic similarity of a unit and a topic as follows: it uses the cosine similarity of the TF-IDF vectors for the unit and the concatenated topic's documents. This corresponds to the semantic interpretation step that is at the core of the well-known ESA model~\cite{gabrilovich:2007}. In Text2vec-SI, the similarity of topics and corpus units is calculated using BERT embeddings~\cite{devlin:2018}. We follow the common approach to generate text embeddings, which is to take the dimension-wise average of the word embeddings for all tokens in the text.\footnote{For efficiency, we limited the embeddings to 10,000~randomly sampled sentences for the topics that had more sentences associated with them.} As a baseline, we implement a direct match approach, which assigns a unit an ontology topic if the topic's text appears in the unit text (ignoring case).

For evaluation, we collect 34,638~pooled query relevance judgments (0.53~inter-annotator agreement as per Krippendorff's~$\alpha$) on 104~randomly selected argument units as queries from 26~corpora. The annotation process is detailed in the appendix.


Based on the similarity scores of the approach, we derive Boolean labels that indicate whether a unit is or is not about one of the ontologies' topics using two policies. The \textit{threshold} policy labels a unit as about a topic if their similarity is above a threshold~$\theta$. The \textit{top}-$k$ policy labels a unit as about a topic if the topic is among the top-$k$ topics with the highest similarity to the unit. We report the parameter of policy with which the approach achieved the highest F$_1$-score on the pooled judgments.

Table~\ref{table-topic-stats-and-classification} shows the results of this evaluation. The baseline produces different results across ontologies---it performs poorly for both the abstract topics in Wikipedia Level~1 and the specific topics in WEF Level~2. The semantic interpretation approach clearly outperforms the baseline for all ontologies in terms of the F$_1$-score. The Text2vec approach outperforms the baseline and semantic interpretation on the most abstract topics (Wikipedia Level~1), but its performance is supbar to that of semantic interpretation on the other ontology levels.

%% file: table-topic-stats-and-classification.tex

\begin{table*}
\centering
\scriptsize
\setlength{\tabcolsep}{1.8pt}

\begin{tabular}{@{}l@{\hspace{3\tabcolsep}}cc@{\hspace{2\tabcolsep}}r@{\hspace{2\tabcolsep}}c@{\hspace{5\tabcolsep}}c@{\hspace{3\tabcolsep}}cccc@{\hspace{5\tabcolsep}}ccc@{\hspace{3\tabcolsep}}lccc@{\hspace{3\tabcolsep}}lccc@{}}

\toprule
\bfseries Ontology & \multicolumn{4}{@{}c@{\hspace{8\tabcolsep}}}{{\bfseries Acquired ontologies} (Section~\ref{topic-ontologies})} & \multicolumn{5}{@{}c@{\hspace{8\tabcolsep}}}{{\bfseries Topic coverage} (Section~\ref{topic-coverage-corpus-topic-labels})} & \multicolumn{10}{@{}c@{}}{{\bfseries Unit categorization} (Section~\ref{topic-coverage-unit-categorization})} \\

\cmidrule(r{8\tabcolsep}){2-5}\cmidrule(l{-2.5\tabcolsep}r{8\tabcolsep}){6-10}\cmidrule{11-21}

& Topics & \multicolumn{3}{@{}c@{\hspace{8\tabcolsep}}}{Topic statistics} && \multicolumn{4}{@{}c@{\hspace{8\tabcolsep}}}{Covered ontology topics}  & \multicolumn{3}{@{}c@{\hspace{4\tabcolsep}}}{Direct match} & \multicolumn{4}{@{}c@{}}{Semantic interpretation} & \multicolumn{4}{@{}c@{\hspace{4\tabcolsep}}}{Text2vec-SI}\\

\cmidrule(l{\tabcolsep}r{8\tabcolsep}){3-5}\cmidrule(r{8\tabcolsep}){7-10}\cmidrule(r{4\tabcolsep}){11-13}\cmidrule(r{4\tabcolsep}){14-17}\cmidrule{18-21}

& & Authors & Docs & Tokens & \makebox(0,0)[b]{\shortstack{Mapped\\[-0.5ex]{}topic\\[-0.5ex]{}labels}} & All & Min & Mean & Max & P & R & F & Policy & P & R & F & Policy & P & R & F \\

\midrule
WEF L1 &           137 & \phantom{0,0}334.1 & 940.7 &           490,576.6 & 1,239 &  87 & 1 & 1.48 & \phantom{0}5 & 0.38 & 0.23 & 0.29 & ${k=12}$ & 0.22 & 0.75 & \textbf{0.34} & $k = 7$ & 0.19 & 0.53 & 0.28\\
WEF L2 &           822 & \phantom{0,0}216.8 & 550.3 &           310,229.7 & \phantom{0}355 &  77 & 1 & 1.32 & \phantom{0}4 & 0.59 & 0.11 & 0.19 & ${k=33}$ & 0.21 & 0.70 & \textbf{0.33} & $\theta = 0.93$ & 0.15 & 0.49 & 0.23\\
WP L1  & \phantom{0}14 &           78,013.7 &  68.0 &           339,088.0 & 1,539 & 14 & 1 & 1.24 & \phantom{0}3 & 0.12 & 0.04 & 0.06 & ${k=3}$ & 0.32 & 0.65 &  0.43 & $k = 2$ & 0.41 & 0.55 & \textbf{0.47}\\
WP L2  &           748 & \phantom{0}1,929.5 &   1.0 & \phantom{00}6,149.1 & 1,453 & 285  & 1 & 1.76 & 16 & 0.47 & 0.34 & 0.40 & ${\theta=0.05}$ & 0.54 & 0.64 & \textbf{0.59} & $\theta = 0.89$ & 0.22 & 0.52 & 0.31\\
DP     & \phantom{0}89 & \phantom{0,0}145.0 &  61.7 & \phantom{0}84,787.6 & 2,002 &  87 & 1 & 2.80 & 10 & 0.49 & 0.37 & 0.42 & ${\theta=0.02}$ & 0.52 & 0.61 & \textbf{0.56} & $k=23$ & 0.36 & 0.80 & 0.50\\

\bottomrule
\end{tabular}

\caption{Statistics for each topic ontology level: for topics and topic documents (Section~\ref{topic-ontologies}), Count of mapped topic labels of the analyzed corpora for each ontology level, Count of all covered ontology topics by the topic labels and the min, max, and mean count of covered ontology topics per topic label (Section~\ref{topic-coverage-corpus-topic-labels}), and the effectiveness of the approaches and baseline in unit categorization (in terms of precision, recall, and  F$_1$-score) (Section~\ref{topic-coverage-unit-categorization}).} 
\label{table-topic-stats-and-classification}
\end{table*}

%% file: eacl23-topic-ontologies-for-argumentation-sum.tex
\section{Conclusion}

The computational argumentation community faces a dilemma: Either carry on as before and risk topic bias, or go the extra mile to ensure topic representativity and create extra work for corpus authors. The latter option is further complicated by the fact that the space of controversial topics is not well explored to date, and that there are no widely accepted argument topic ontologies as of yet. In this paper, we give a glimpse into a future in which the argument topic space has been mapped and is accessible to corpus construction and designing experiments: We identified three authoritative sources of ontologic knowledge with respect to argument topics. For each ontology, we reveal the topic coverage of 31 argument corpora that are provided with topic labels by aligning the topic labels of a corpus to the ontologies topics. To assess the topic coverage of non-labeled corpora, we introduce an approach that identifies the ontology topics of an argumentative text, reaching an F$_1$ of~0.59.

Our analyses show that the topic coverage of the studied argument corpora is both limited to only a subset of the ontologies topics and skewed. The majority of topics that require domain knowledge such as those on mental health, philosophy, or international security are marginally covered in the analyzed argument corpora. This renders existing argumentation technologies more suitable to teach people how to construct arguments than to support them taking decisions about complex topics. For a mature development of argumentation technology, a careful sampling and controlling of the topics should be employed while constructing corpora, designing experiments, or applying classifiers. 

In future work, major tasks are to further explore the argument topic space regarding matters not yet covered by the existing topic ontologies and to unify the different ontologies. Besides ``is part of''-relations between topics, other relation types may be considered as well, thereby inducing a topic knowledge base. However, already the work presented here can assist in selecting arguments for corpus construction and model training.

%% file: eacl23-topic-ontologies-for-argumentation-limitations.tex
\section{Limitations}

The three topic ontologies which we used to assess the topic coverage of argument corpora come from recognized sources and cover different domains. Nevertheless, these topic ontologies might not fully cover all possible controversial topics that are relevant to argumentation (e.g., those topics related to private life). Having said that, we do believe that our paper sets a cornerstone for studying topic bias in argument corpora, which researchers can extend.

Another limitation of this study is the moderate effectiveness achieved by our approaches for unit categorization. Our approaches for unit categorization achieved moderate effectiveness because of the large space of controversial topics (about 742 for Wikipedia). Future research can improve upon our approach by utilizing the structure of the topic ontology using hierarchical classifiers. Hierarchical classifiers first map a document to one topic in the upper level and then consider only the subtopics of this topic for classification in the lower levels. In this way, the space of controversial topics in the lower levels can be largely reduced.

%% file: eacl23-topic-ontologies-for-argumentation-supplementary.tex
\section{Appendix}
\subsection{Corpus Topic Labels Mapping to Level~2 Topics}

\begin{figure*}[t]
\includegraphics[width=\linewidth]{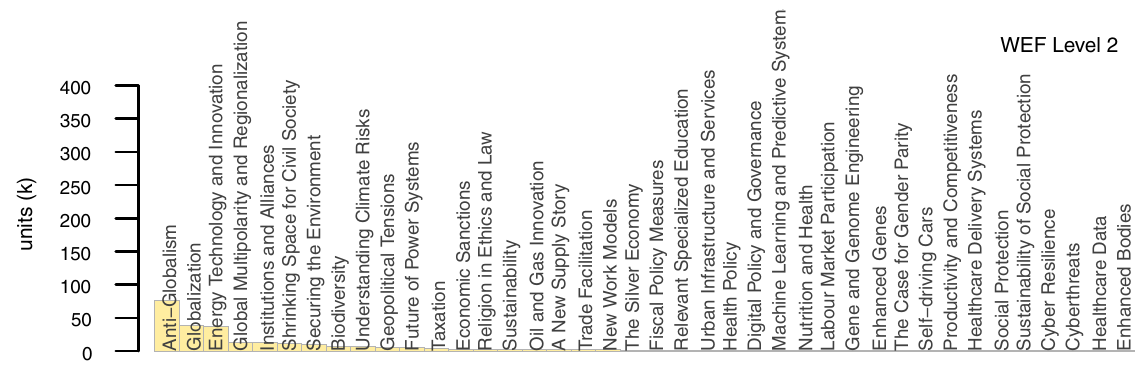}\\[1ex]
\includegraphics[width=\linewidth]{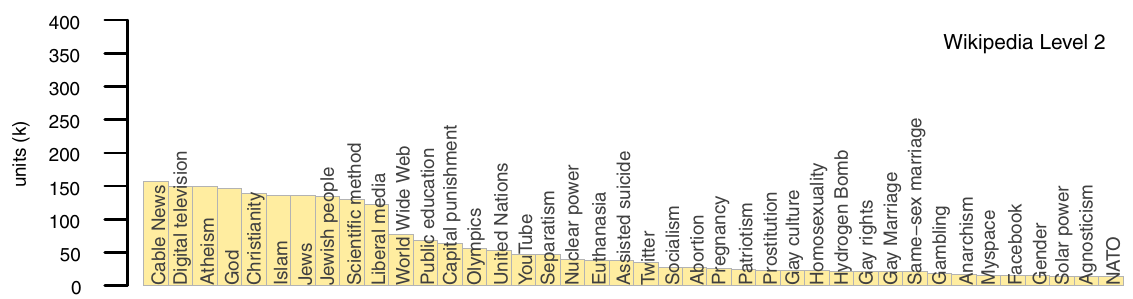}
\caption{Omitted graphs from Figure~2, Section~5.3}
\label{figure2-supplemental}
\end{figure*}

For completeness, we Figure~\ref{figure2-supplemental} show the two graphs that are omitted from Figure~2 of the paper as their fine-grained topics are less relevant for the discussion in Section~5.3.

\subsection{Annotation Procedure for Unit Categorization}

\begin{figure*}[t]
\includegraphics[width=\linewidth]{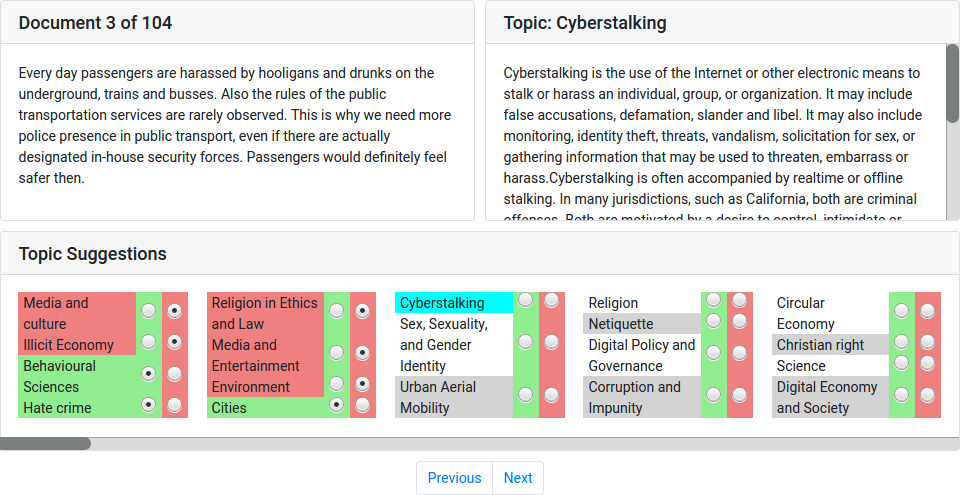}
\caption{Assessment interface for topic labeling.}
\label{assessment-interface}
\end{figure*}

In order to assess the effectiveness of the approaches and baseline outlined in the paper, we employ a pooled evaluation, as it is standard for information retrieval evaluations, where there are too many instances for a complete manual annotation. We randomly sampled four units from 26~corpora, which were all annotated by three expert annotators. The annotators were instructed to label a topic as about the unit if they could imagine a discussion on the topic for which the unit would be relevant. For each unit, we annotated for aboutness only those topics which are among the five topics with the highest similarity to this unit according to at least one of the approaches. The employed assessment interface (see Figure~\ref{assessment-interface}) shows the unit (top left), the current topic (top right), as well as all topics in the pool for that unit (bottom; the current topic is marked blue, whereas already annotated topics are marked green (about) and red (not about). The same interface has been used for the topic label annotations.

To reduce biases, both the units and the topics were shown in a different and random order to each assessor. The annotation took about 40~hours. The annotation process resulted in an inter-annotator agreement of 0.53 in terms of Krippendorff's~$\alpha$ and produced a total of 34,638~annotations of topic-unit pairs, about~2\% of what would have been needed for a complete annotation.


